\documentclass[lettersize,journal]{IEEEtran}

\usepackage{amsmath,amsfonts}
\usepackage{algorithmic}
\usepackage{algorithm}

\usepackage{array}
\usepackage{multirow}
\usepackage{graphicx}
\usepackage{subfig}
\usepackage{adjustbox}

\usepackage{textcomp}
\usepackage{stfloats}
\usepackage{url}
\usepackage{verbatim}
\usepackage{changepage} 
\usepackage{cite}
\usepackage{xcolor}
\usepackage[outline]{contour}
\usepackage{hyperref}
\usepackage{orcidlink}
\usepackage{tcolorbox}

\hypersetup{
    colorlinks=false,
    pdfborder={0 0 0},
}

\begin{document}

\title{Demystifying ChatGPT: How It Masters Genre Recognition}

\author{
    Subham Raj\,\orcidlink{0000-0002-2289-6771} \and
    Sriparna Saha\,\orcidlink{0000-0001-5458-9381} \and
    Brijraj Singh\,\orcidlink{0000-0003-0626-7905} \and
    Niranjan Pedanekar\,\orcidlink{0009-0009-5381-5450}
    
    \thanks{Subham Raj and Sriparna Saha are with the Department of Computer Science and Engineering, Indian Institute of Technology Patna, India (emails: \texttt{subham\_2221cs25@iitp.ac.in}; \texttt{sriparna@iitp.ac.in}).}
    
    \thanks{Brijraj Singh and Niranjan Pedanekar are with Sony Research India (emails: \texttt{Brijraj.Singh@sony.com}; \texttt{Niranjan.Pedanekar@sony.com}).}
}

\maketitle

\begin{abstract}
The introduction of ChatGPT has garnered significant attention within the NLP community and beyond. Previous studies have demonstrated ChatGPT's substantial advancements across various downstream NLP tasks, highlighting its adaptability and potential to revolutionize language-related applications. However, its capabilities and limitations in genre prediction remain unclear. This work analyzes three Large Language Models (LLMs) using the MovieLens-100K dataset to assess their genre prediction capabilities. Our findings show that ChatGPT, without fine-tuning, outperformed other LLMs, and fine-tuned ChatGPT performed best overall. We set up zero-shot and few-shot prompts using audio transcripts/subtitles from movie trailers in the MovieLens-100K dataset, covering 1682 movies of 18 genres, where each movie can have multiple genres. Additionally, we extended our study by extracting IMDb movie posters to utilize a Vision Language Model (VLM) with prompts for poster information. This fine-grained information was used to enhance existing LLM prompts. In conclusion, our study reveals ChatGPT's remarkable genre prediction capabilities, surpassing other language models. The integration of VLM further enhances our findings, showcasing ChatGPT's potential for content-related applications by incorporating visual information from movie posters.
\end{abstract}


\section{Introduction}
\IEEEPARstart{I}{n} recent years, large language models (LLMs) have emerged as a transformative tool in natural language processing (NLP) research, revolutionizing various fields ranging from machine translation\cite{brants2007large,wang2023document,moslem2023adaptive} and text generation\cite{zhang2022survey,mccoy2023much} to sentiment analysis\cite{zhang2023instruct,zhang2023enhancing} and information retrieval\cite{zhu2023large}. 
These models, often based on deep learning architectures, possess the capacity to process and comprehend vast amounts of textual data, enabling them to learn complex patterns and relationships within language. 
Research papers focusing on large language models have delved into several crucial aspects, including model architecture design, training techniques, fine-tuning strategies, and applications across different domains. 
These studies frequently highlight the challenges associated with model scalability, computational resources, ethical considerations, and the implications of LLMs for society, emphasizing the need for responsible deployment and robust evaluation methodologies.
By continually advancing the capabilities of large language models, researchers are actively contributing to the ongoing evolution of artificial intelligence, paving the way for groundbreaking innovations in language understanding and generation.

Multi-label movie genre classification is a significant task in the field of machine learning and natural language processing. 
This task presents unique challenges compared to traditional single-label classification, as movies often belong to multiple genres simultaneously, leading to complex label dependencies and potential ambiguities.
Researchers have explored various approaches, including deep learning models such as convolutional neural networks (CNNs) \cite{simoes2016movie,allamy20211d}, recurrent neural networks (RNNs)\cite{behrouzi2023multimodal,dai2016long}, and transformer-based architectures like BERT and its variants \cite{akalp2021language,marijic2023predicting} for solving the genre prediction task. 



In recent studies, VLMs have demonstrated remarkable performance in information extraction from images, as evidenced by the notable works of \cite{zhou2022learning,liu2021image,ossowski2024prompting,shen2024multitask}. The decision to incorporate VLMs into our research was inspired by the wealth of hidden information embedded in movie posters, which significantly contributes to predicting a movie's genre. While several deep learning models\cite{barney2019predicting,kundalia2020multi,chu2017movie} have leveraged movie posters for multi-label genre classification, an intriguing observation from recent literature highlights the underexplored potential of VLMs in genre prediction tasks. This underscores a unique opportunity for our study to bridge this gap and potentially enhance genre prediction by harnessing the capabilities of VLMs. For our study, we have explored Llava-7B\cite{liu2023llava} model\footnote{https://huggingface.co/4bit/llava-v1.5-13b-3GB}.

In this work, we have performed the benchmarking of different LLMs in terms of multi-label movie genre classification. As per our knowledge, we are the first one to perform this benchmarking. We have set up a \textbf{zero-shot prompt and few-shot prompt} in order to trigger LLMs to generate a response from it. We utilized the extracted trailer subtitles from the movies within the MovieLens-100K dataset, employing them as a stimulating prompt. In summary, we have the following major contributions/findings:
\begin{itemize}
    \item \textit{Extended the MovieLens-100K dataset by extracting the movie's trailer's subtitles and the movie's poster.}
    \item \textit{ChatGPT(gpt-3.5-turbo) and its fine-tuned version showed consistent advantages in terms of genre prediction compared to other LLMs like text-davinci-002 and text-davinci-003. Detailed descriptions of these LLMs can be found \href{https://platform.openai.com/docs/models}{here}. }
    
    \item \textit{ChatGPT(gpt-3.5-turbo) under zero-shot setting outperformed traditional classifiers.}
    \item \textit{We have performed an interesting study for the genre prediction task by integrating llava VLM and LLM.}
    \item \textit{Cost analysis showed that fine-tuned ChatGPT performed much better compared to non-fine-tuned ChatGPT provided the cost increases, but the few-shot settings performance is at par with the performance of zero-shot settings, hence zero-shot setting is preferred.}
\end{itemize}

We hope that this preliminary evaluation of ChatGPT in genre prediction can provide new perspectives on both assessing the capabilities of LLMs and utilizing LLMs, such as ChatGPT, to enhance works such as recommendation systems as these two tasks are closely related \cite{raj2023multi}. 

The remainder of this paper is as follows. Section 2 discusses the dataset. Section 3 discusses the different LLMs and their prompting techniques for genre prediction. Section 4 discusses the experiment and result, which is followed by a conclusion and future works.

\section{Dataset Overview and its Augmentation}
The MovieLens dataset is a curated collection of movie ratings and user demographic information. It comprises 100,000 movie ratings, represented on a scale from 1 to 5, provided by 943 distinct users for 1,682 movies. It has meta information such as age, gender, occupation, and zip code available for every user, and genre information is available for every movie.

In order to set up a prompt for LLMs, we needed the movie's trailer's subtitles; therefore, we have extended this dataset by incorporating subtitles for each movie's trailer present in the MovieLens-100K dataset. Trailer subtitles are extracted via two sources of transcription. 
Firstly for those movie's trailer's whose subtitles were enabled on youtube, we use \href{https://pypi.org/project/youtube-transcript-api/}{youtube-transcript-api} to transcribe them. For those whose trailer subtitles were disabled on YouTube, we use \href{https://huggingface.co/spaces/jeffistyping/Youtube-Whisperer}{Youtube-Whisperer} to transcribe.

The movie's poster was needed to trigger VLM. In order to collect it, we scraped the IMDb website via \href{https://beautiful-soup-4.readthedocs.io/en/latest/}{BeautifulSoup}. This results in the collection of 1682 movie posters present in the dataset. These visuals played a crucial role in expanding our exploration of genre prediction, injecting a cinematic touch to our analysis.

\section{Benchmark Setup}


\subsection{Prompting Techniques}

\subsubsection{Zero-shot prompt}
In our study for the task of movie genre identification, we first employ a zero-shot prompting approach with large language models. The prompt we utilize consists of a comprehensive list of movie genres, denoted as $G$. Additionally, this prompt incorporates the movie's trailer's subtitles, designated as $S$. Given $G$ in square brackets: 
\newline
\textit{$G = [$Action, Adventure, Animation, Children's, Comedy, Crime, Documentary, Drama, Fantasy, Film-Noir, Horror, Musical, Mystery, Romance, Sci-Fi, Thriller, War, Western$]$}

we aim to predict the potential genres $G'$ of a movie, represented as 
\begin{equation}
    G' = \textit{LLM}(G, S)
\end{equation}
where $G'$ $\subseteq$ $G$.

Figure \ref{example_3}(a) shows the LLM zero-shot prompt.
\subsubsection{Few-shot prompt}
Under a few show settings, we aim to predict the potential genres $G'$ of a movie, represented as 
\begin{equation}
    G' = \textit{LLM}(G, S | S')
\end{equation}
where $G'$ $\subseteq$ $G$ and $S'$ denote the set of examples provided having subtitles as an input and genre as an output for a movie. For experimental purposes, we have used two reference examples. The description of the prompt is provided in Figure \ref{example_3}(b). 

\subsection{Fine-tuning Instruction}
We adopt a conversational chat format of instructions to fine-tune the language model specifically for the task of movie genre identification. The conversation is defined as a sequence of messages, denoted as 
\begin{equation}
    P = \{(R_i, C_i) | i = 1,..3\}
\end{equation}

\begin{figure*}
\begin{center}
\subfloat[Zero-shot prompt for performing LLM inference]{\includegraphics[width = 3in,height=2in]{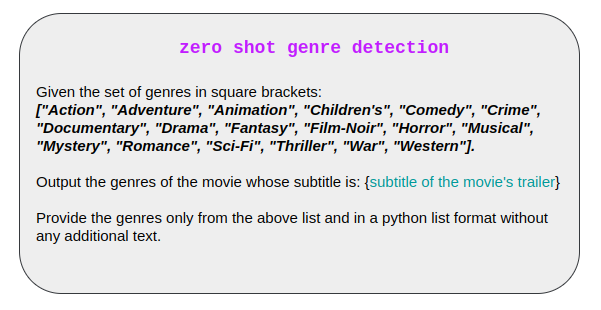}} 
\quad
\subfloat[Few-shot prompt for performing LLM inference]{\includegraphics[width = 3in,height=2in]{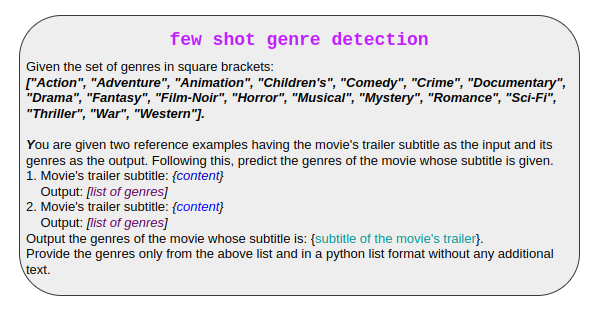}}
\\
\subfloat[Instruction and prompt for Fine-Tuning]{\includegraphics[width = 4in,height=3in]{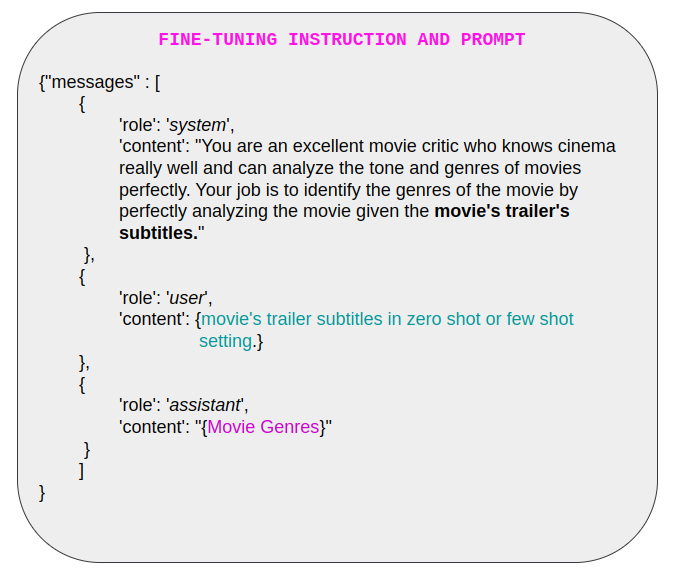}}
\end{center}
\caption{LLM Prompt and Instruction}
\label{example_3}
\end{figure*}

Here, $R_i$ indicates the role of the $i$th message, and $C_i$ represents the content of the $i$th message. The user message, which incorporates the trailer subtitles, serves as the model's input, while the assistant message contains the model's predictions for the movie genres. This approach allows us to effectively harness the power of LLMs by fine-tuning them for the specific task of identifying movie genres with high accuracy. Figure \ref{example_3}(c) shows the instruction and prompt for fine-tuning. This fine-tuning is done for both the zero-shot and few-shot settings.

\section{Experiments and Result}

We conducted experiments to evaluate text-davinci-002, text-davinci-003, and gpt-3.5-turbo(ChatGPT) LLMs to answer the following research question:
\begin{itemize}
    \item \textbf{RQ1:} How does the performance of different LLMs, with and without fine-tuning, vary when evaluated using diverse classification metrics for genre prediction?
    \item \textbf{RQ2:} How does the LLMs-based genre prediction system compare with the traditional classification method?
    \item \textbf{RQ3:} How much is the trade-off between the cost of gpt-3.5-turbo under zero-shot vs few-shot and under fine-tuned vs non-fine-tuned settings?
    \item \textbf{RQ4:} How does the number of prompt shots affect the performance of gpt-3.5-turbo?
\end{itemize}

\subsection {Experimental Setup}
The experiment involved running multi-label movie genre classification on the MovieLens-100K dataset, which has been extensively used in collaborative filtering and recommendation system research. The dataset comprises user movie ratings and demographic information, making it a valuable resource for evaluating the performance of language models in genre prediction.

\subsubsection{Hyperparameter settings}
In the utilization of the Language Model (LLM), key parameters such as \textbf{temperature=0.1, top\_p=1, and max\_tokens=270} play a pivotal role in shaping the output and ensuring optimal performance. These hyperparameters influence the generation process, controlling factors like randomness, sequence length, and the probability distribution of the generated tokens.

To fine-tune the gpt-3.5-turbo model, we employed an 80:20 split for the MovieLens-100K. 80\% of the dataset, specifically 1,205 movies, for fine-tuning/training. The remaining 20\%, or 302 movies, were for testing the model.


\subsection {RQ1: Overall Performance}
\begin{figure*}
\subfloat[Radar plot for Precision of LLMs]{\includegraphics[width = 3in]{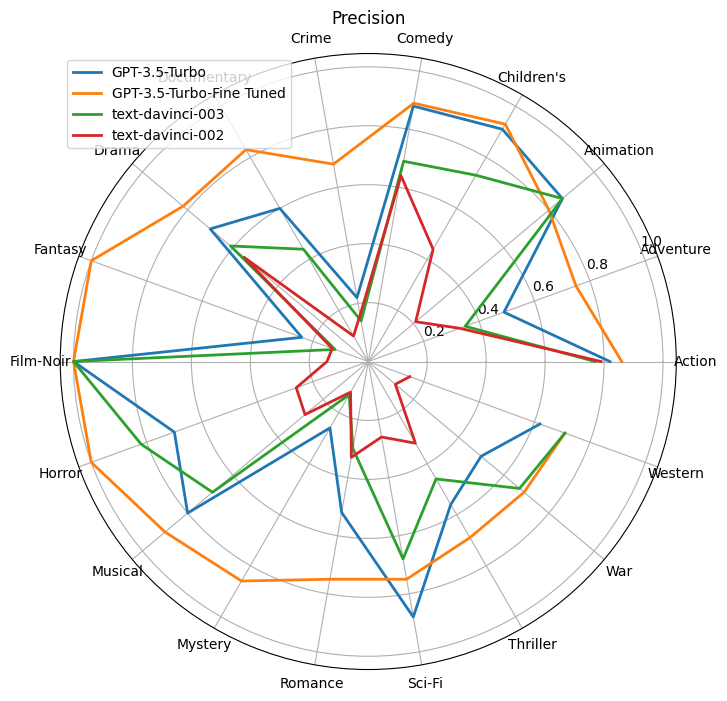}} 
\quad
\subfloat[Radar plot for Recall of LLMs]{\includegraphics[width = 3in]{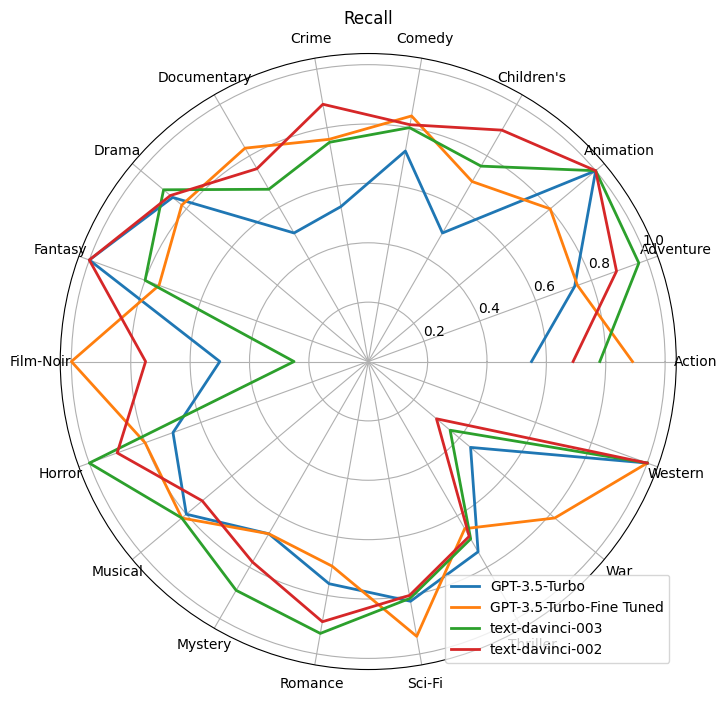}}
\\
\centering\subfloat[Radar plot for F1-score of LLMs]{\includegraphics[width = 3in]{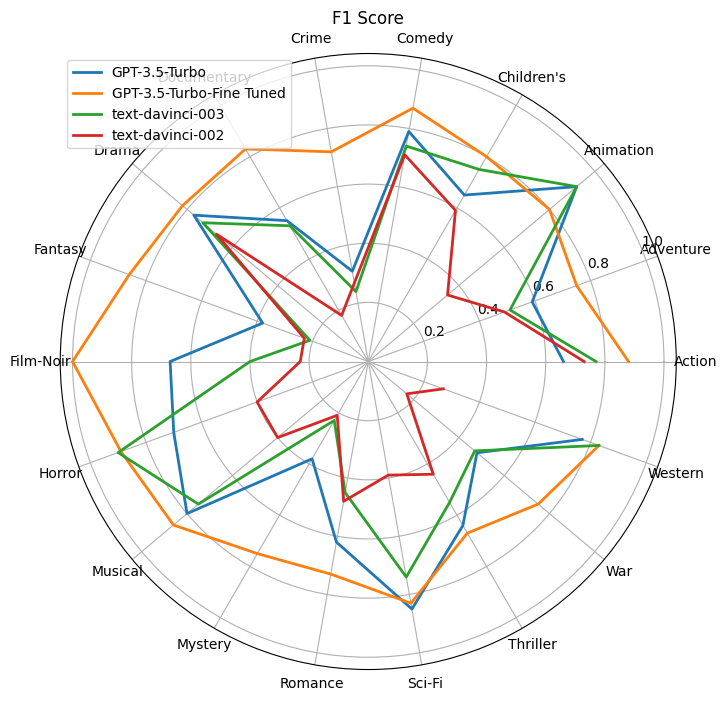}}
\caption{Performance of LLMs under zero-shot settings on Movielens-100K dataset. }
\label{example_1}
\end{figure*}

\begin{figure*}
\subfloat[Heatmap for Precision of LLMs]{\includegraphics[width = 3.5in,height=1.7in]{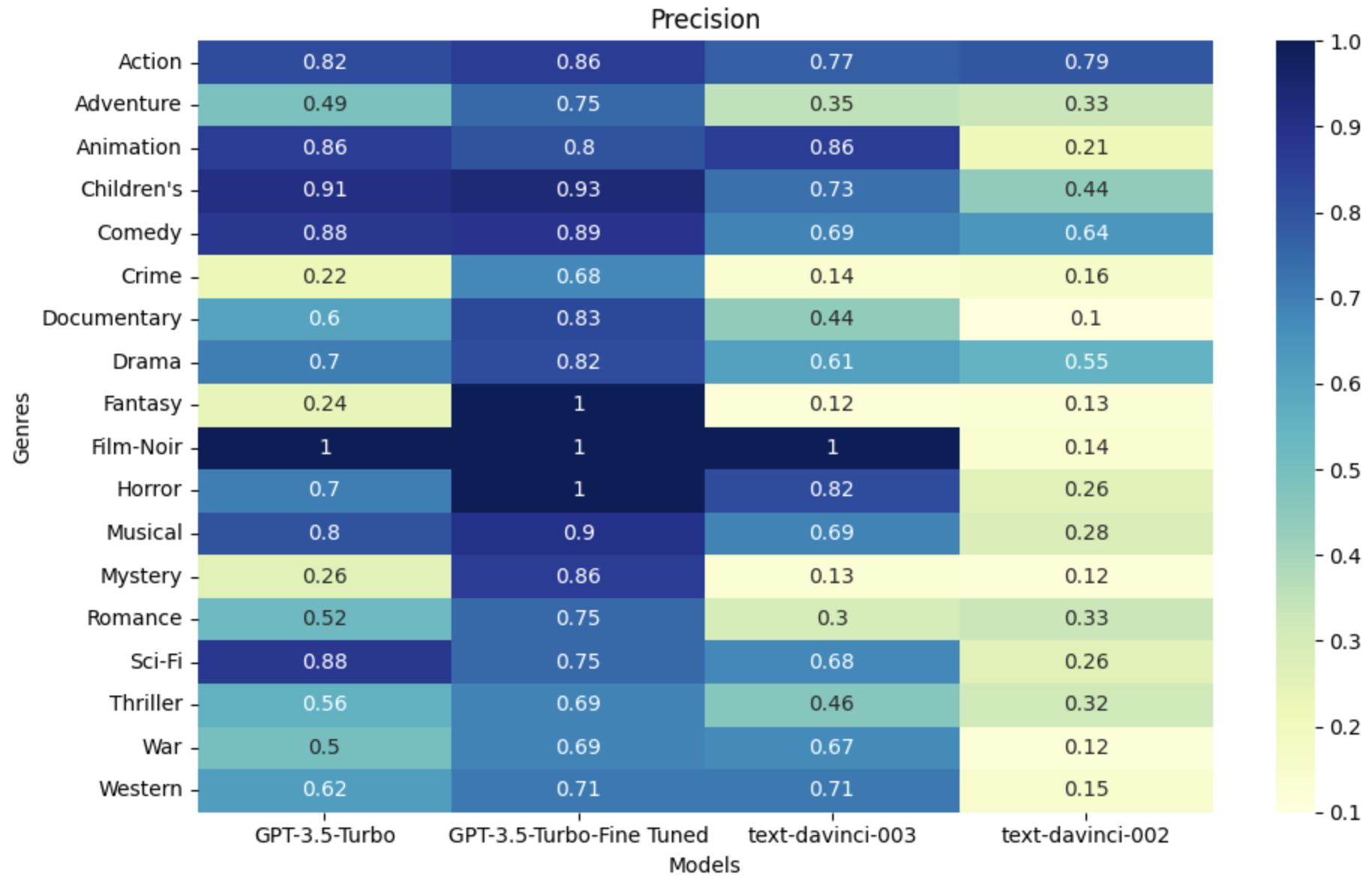}} 
\subfloat[Heatmap for Recall of LLMs]{\includegraphics[width = 3.5in,height=1.7in]{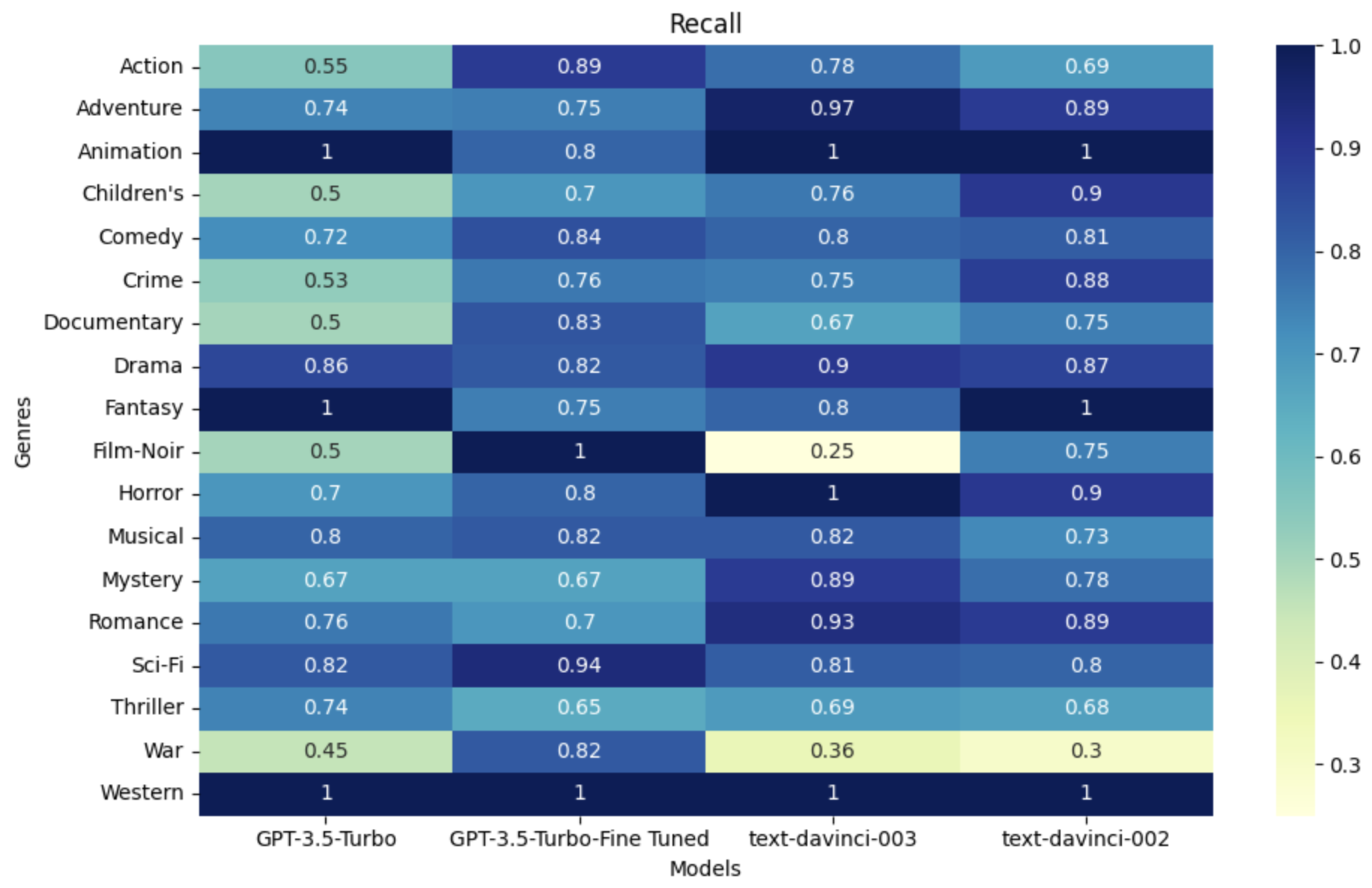}}\\
\centering\subfloat[Heatmap for F1-score of LLMs]{\includegraphics[width = 3.5in,height=1.7in]{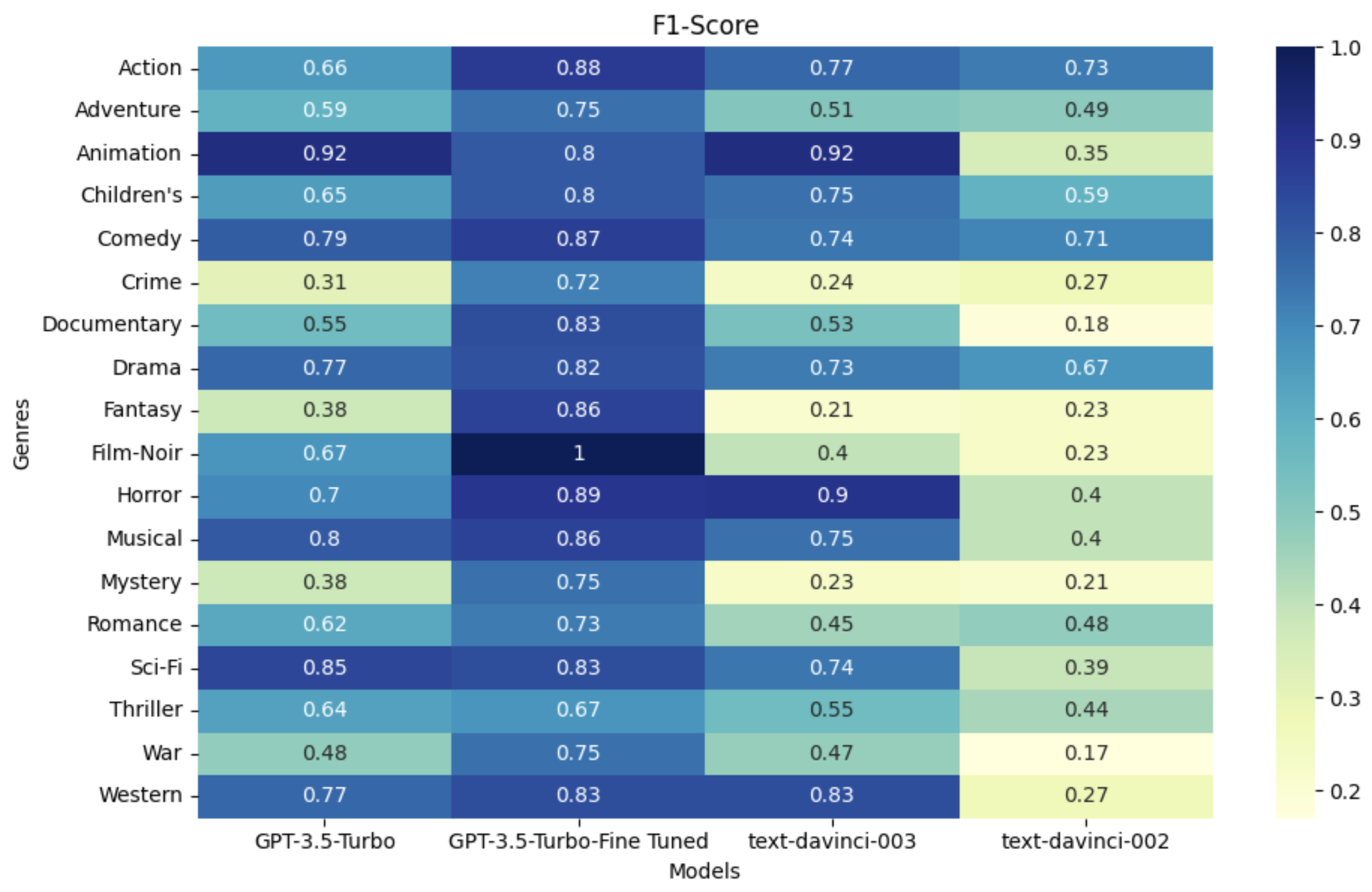}}
\caption{Performance of LLMs under zero-shot settings on Movielens-100K dataset. The Y-axis represents different genre labels, and the X represents different models.}
\label{example}
\end{figure*}

Table \ref{ablation_ML} and \ref{ablation_ML_1} shows the performance of four large language models on the Movielens-100K dataset, across 18 distinct movie genres under zero-shot settings. 
Figure \ref{example_1} and \ref{example} also illustrate the plot for the performance of different LLMs under zero-shot settings in the form of a radar plot of precision, recall, and F1-score and in the form of a heatmap of precision, recall, and F1-score respectively. We have extended our experiments to few-shot settings whose results are illustrated in the Table \ref{ablation_ML_2} and \ref{ablation_ML_3}. 
Following are the observations and conclusions regarding it:

\textbf{Fine-tuned gpt-3.5-turbo and gpt-3.5-turbo performed much better than any random classifier in terms of genre prediction in all the metrics.} Specifically, all three LLMs have performed better than random classifiers. In terms of genres such as action, adventure, children's, comedy, crime, horror, musical, and war, fine-tuned gpt-3.5-turbo has shown remarkable performance. Additionally, most answers of LLMs are compliant due to the capability of in-context learning. These results showcase that LLMs have the potential to facilitate genre prediction systems.

\textbf{In comparison to text-davinci-002 and text-davinci-003, ChatGPT has shown better performance in terms of all metrics.} ChatGPT has outperformed other LLMs in 54 out of 54 instances, including a few times where the results were idential. The exceptional performance of ChatGPT can be attributed to the fact that it has an exceptional capacity for language understanding and reasoning. 

\textbf{Compared to ChatGPT, fine-tuned ChatGPT under zero-shot or few-shot has proven to be better because it got trained properly on the samples.} In comparison to the non-fine-tuned version, this model showed higher recall for several genres, including Action, Adventure, Children's, and Sci-Fi, which suggests that the non-fine-tuned version may have missed some relevant instances. However, ChatGPT has performed better for children's and romance genres compared to fine-tuned ChatGPT, which may be the case of underfitting due to limited samples. Overall, fine-tuned ChatGPT under few shot settings has outperformed all other settings of different LLMs.

\begin{table*}

\scriptsize
\centering
\renewcommand{\arraystretch}{1.5}
\begin{tabular}{|c|c|c|c|c|c|c|c|c|c|c|c|c|}
\hline
\multirow{2}{*}{\textbf{Genre}} & \multicolumn{3}{c|}{\textbf{text-davinci-002}} & \multicolumn{3}{c|}{\textbf{text-davinci-003}} & \multicolumn{3}{c|}{\textbf{gpt-3.5-turbo}} & \multicolumn{3}{c|}{\textbf{gpt-3.5-turbo-fine-tuned}}\\
     & Precision&Recall&F1-score& Precision&Recall&F1-score & Precision&Recall&F1-score & Precision&Recall&F1-score\\
\hline
Action&0.79&0.69&0.73&0.77&0.78&0.77&0.82&0.55&0.66&\textbf{0.86}\contour{green}{($\Uparrow$}\contour{green}{5$\%$)}&\textbf{0.89}\contour{green}{($\Uparrow$}\contour{green}{62$\%$)}&\textbf{0.88}\contour{green}{($\Uparrow$}\contour{green}{33$\%$)}\\ 
\hline
Adventure&0.33&0.89&0.49&0.35&0.97&0.51&0.49&0.74&0.59&\textbf{0.75}\contour{green}{($\Uparrow$}\contour{green}{53$\%$)}&\textbf{0.75}\contour{green}{($\Uparrow$}\contour{green}{2$\%$)}&\textbf{0.75}\contour{green}{($\Uparrow$}\contour{green}{27$\%$)}\\ 
\hline
Animation&0.21&1.00&0.35&0.86&1.00&0.92&\textbf{0.86}&\textbf{1.00}&\textbf{0.92}&0.80\contour{red}{($\Downarrow$}\contour{red}{7$\%$)}&0.80\contour{red}{($\Downarrow$}\contour{red}{20$\%$)}&0.80\contour{red}{($\Downarrow$}\contour{red}{13$\%$)}\\ 
\hline
Children's&0.44&0.90&0.59&0.73&0.76&0.75&0.91&0.50&0.65&\textbf{0.93}\contour{green}{($\Uparrow$}\contour{green}{2$\%$)}&\textbf{0.70}\contour{green}{($\Uparrow$}\contour{green}{40$\%$)}&\textbf{0.80}\contour{green}{($\Uparrow$}\contour{green}{23$\%$)}\\ 
\hline
Comedy&0.64&0.81&0.71&0.69&0.80&0.74&0.88&0.72&0.79&\textbf{0.89}\contour{green}{($\Uparrow$}\contour{green}{1$\%$)}&\textbf{0.84}\contour{green}{($\Uparrow$}\contour{green}{17$\%$)}&\textbf{0.87}\contour{green}{($\Uparrow$}\contour{green}{10$\%$)}\\ 
\hline
Crime&0.16&0.88&0.27&0.14&0.75&0.24&0.22&0.53&0.31&\textbf{0.68}\contour{green}{($\Uparrow$}\contour{green}{209$\%$)}&\textbf{0.76}\contour{green}{($\Uparrow$}\contour{green}{43$\%$)}&\textbf{0.72}\contour{green}{($\Uparrow$}\contour{green}{132$\%$)}\\ 
\hline
Documentary&0.10&0.75&0.18&0.44&0.67&0.53&0.60&0.50&0.55&\textbf{0.83}\contour{green}{($\Uparrow$}\contour{green}{38$\%$)}&\textbf{0.83}\contour{green}{($\Uparrow$}\contour{green}{66$\%$)}&\textbf{0.83}\contour{green}{($\Uparrow$}\contour{green}{51$\%$)}\\ 
\hline
Drama&0.55&0.87&0.67&0.61&0.90&0.73&0.70&\textbf{0.86}&0.77&\textbf{0.82}\contour{green}{($\Uparrow$}\contour{green}{17$\%$)}&0.82\contour{red}{($\Downarrow$}\contour{red}{5$\%$)}&\textbf{0.82}\contour{green}{($\Uparrow$}\contour{green}{6$\%$)}\\ 
\hline
Fantasy&0.13&1.00&0.23&0.12&0.80&0.21&0.24&\textbf{1.00}&0.38&\textbf{1.00}\contour{green}{($\Uparrow$}\contour{green}{317$\%$)}&0.75\contour{red}{($\Downarrow$}\contour{red}{25$\%$)}&\textbf{0.86}\contour{green}{($\Uparrow$}\contour{green}{126$\%$)}\\ 
\hline
Film-Noir&0.14&0.75&0.23&1.00&0.25&0.40&\textbf{1.00}&0.50&0.67&\textbf{1.00}\contour{blue}{($=$}\contour{blue}{0$\%$)}&\textbf{1.00}\contour{green}{($\Uparrow$}\contour{green}{100$\%$)}&\textbf{1.00}\contour{green}{($\Uparrow$}\contour{green}{49$\%$)}\\ 
\hline
Horror&0.26&0.90&0.40&0.82&1.00&0.90&0.70&0.70&0.70&\textbf{1.00}\contour{green}{($\Uparrow$}\contour{green}{43$\%$)}&\textbf{0.80}\contour{green}{($\Uparrow$}\contour{green}{14$\%$)}&\textbf{0.89}\contour{green}{($\Uparrow$}\contour{green}{27$\%$)}\\ 
\hline
Musical&0.28&0.73&0.40&0.69&0.82&0.75&0.80&0.80&0.80&\textbf{0.90}\contour{green}{($\Uparrow$}\contour{green}{13$\%$)}&\textbf{0.82}\contour{green}{($\Uparrow$}\contour{green}{3$\%$)}&\textbf{0.86}\contour{green}{($\Uparrow$}\contour{green}{8$\%$)}\\ 
\hline
Mystery&0.12&0.78&0.21&0.13&0.89&0.23&0.26&\textbf{0.67}&0.38&\textbf{0.86}\contour{green}{($\Uparrow$}\contour{green}{231$\%$)}&\textbf{0.67}\contour{blue}{($=$}\contour{blue}{0$\%$)}&\textbf{0.75}\contour{green}{($\Uparrow$}\contour{green}{97$\%$)}\\ 
\hline
Romance&0.33&0.89&0.48&0.30&0.93&0.45&0.52&\textbf{0.76}&0.62&\textbf{0.75}\contour{green}{($\Uparrow$}\contour{green}{44$\%$)}&0.70\contour{red}{($\Downarrow$}\contour{red}{8$\%$)}&\textbf{0.73}\contour{green}{($\Uparrow$}\contour{green}{18$\%$)}\\ 
\hline
Sci-Fi&0.26&0.80&0.39&0.68&0.81&0.74&\textbf{0.88}&0.82&\textbf{0.85}&0.75\contour{red}{($\Downarrow$}\contour{red}{15$\%$)}&\textbf{0.94}\contour{green}{($\Uparrow$}\contour{green}{15$\%$)}&0.83\contour{red}{($\Downarrow$}\contour{red}{2$\%$)}\\ 
\hline
Thriller&0.32&0.68&0.44&0.46&0.69&0.55&0.56&\textbf{0.74}&0.64&\textbf{0.69}\contour{green}{($\Uparrow$}\contour{green}{23$\%$)}&0.65\contour{red}{($\Downarrow$}\contour{red}{12$\%$)}&\textbf{0.67}\contour{green}{($\Uparrow$}\contour{green}{5$\%$)}\\ 
\hline
War&0.12&0.30&0.17&0.67&0.36&0.47&0.50&0.45&0.48&\textbf{0.69}\contour{green}{($\Uparrow$}\contour{green}{38$\%$)}&\textbf{0.82}\contour{green}{($\Uparrow$}\contour{green}{82$\%$)}&\textbf{0.75}\contour{green}{($\Uparrow$}\contour{green}{56$\%$)}\\ 
\hline
Western&0.15&1.00&0.27&0.71&1.00&0.83&0.62&\textbf{1.00}&0.77&\textbf{0.71}\contour{green}{($\Uparrow$}\contour{green}{15$\%$)}&\textbf{1.00}\contour{blue}{($=$}\contour{blue}{0$\%$)}&\textbf{0.83}\contour{green}{($\Uparrow$}\contour{green}{8$\%$)}\\ 
\hline

\end{tabular}
\caption{Performance of four large language models on the Movielens-100K dataset, across 18 distinct movie genres under zero-shot settings. The percentage variations in Precision, Recall, and F1-Score metrics, as illustrated in the fine-tuned version of ChatGPT, are compared to the baseline ChatGPT (gpt-3.5-turbo).}\label{ablation_ML}
\end{table*}

\begin{table*}
\scriptsize
\centering
\renewcommand{\arraystretch}{1.5}
\begin{tabular}{|c|c|c|c|c|c|c|c|c|c|c|c|c|}
\hline
\multirow{2}{*}{\textbf{Average Metrics}} & \multicolumn{3}{c|}{\textbf{text-davinci-002}} & \multicolumn{3}{c|}{\textbf{text-davinci-003}} & \multicolumn{3}{c|}{\textbf{gpt-3.5-turbo}} & \multicolumn{3}{c|}{\textbf{gpt-3.5-turbo-fine-tuned}}\\
     & Precision&Recall&F1-score& Precision&Recall&F1-score & Precision&Recall&F1-score & Precision&Recall&F1-score\\
\hline
Micro Avg&0.35&0.81&0.49&0.46&0.82&0.59&0.62&0.73&0.67&\textbf{0.82}\contour{green}{($\Uparrow$}\contour{green}{32$\%$)}&\textbf{0.80}\contour{green}{($\Uparrow$}\contour{green}{10$\%$)}&\textbf{0.81}\contour{green}{($\Uparrow$}\contour{green}{31$\%$)}\\ 
\hline
Macro Avg&0.30&0.81&0.49&0.56&0.79&0.60&0.64&0.71&0.64&\textbf{0.83}\contour{green}{($\Uparrow$}\contour{green}{30$\%$)}&\textbf{0.81}\contour{green}{($\Uparrow$}\contour{green}{14$\%$)}&\textbf{0.81}\contour{green}{($\Uparrow$}\contour{green}{27$\%$)}\\ 
\hline
Weighted Avg&0.46&0.81&0.56&0.58&0.82&0.65&0.69&0.73&0.69&\textbf{0.82}\contour{green}{($\Uparrow$}\contour{green}{19$\%$)}&\textbf{0.80}\contour{green}{($\Uparrow$}\contour{green}{10$\%$)}&\textbf{0.81}\contour{green}{($\Uparrow$}\contour{green}{18$\%$)}\\ 
\hline
Samples Avg&0.47&0.83&0.55&0.48&0.84&0.58&0.66&0.75&0.67&\textbf{0.85}\contour{green}{($\Uparrow$}\contour{green}{29$\%$)}&\textbf{0.82}\contour{green}{($\Uparrow$}\contour{green}{9$\%$)}&\textbf{0.81}\contour{green}{($\Uparrow$}\contour{green}{21$\%$)}\\ 
\hline
\end{tabular}
\caption{Performance of four large language models on the Movielens-100K dataset under zero-shot settings, across average metrics. The percentage variations in Precision, Recall, and F1-Score metrics, as illustrated in the fine-tuned version of ChatGPT, are compared to the baseline ChatGPT (gpt-3.5-turbo).}\label{ablation_ML_1}
\end{table*}

\begin{table}


\centering
\renewcommand{\arraystretch}{1.5}
\resizebox{\columnwidth}{!}{%
\begin{tabular}{|c|c|c|c|c|c|c|}
\hline
\multirow{2}{*}{\textbf{Genre}}  & \multicolumn{3}{c|}{\textbf{gpt-3.5-turbo}} & \multicolumn{3}{c|}{\textbf{gpt-3.5-turbo-fine-tuned}}\\
     & Precision&Recall&F1-score& Precision&Recall&F1-score\\
\hline
Action&0.84&0.55&0.67&\textbf{0.86}\contour{green}{($\Uparrow$}\contour{green}{2$\%$)}&\textbf{0.89}\contour{green}{($\Uparrow$}\contour{green}{62$\%$)}&\textbf{0.88}\contour{green}{($\Uparrow$}\contour{green}{31$\%$)}\\ 
\hline
Adventure&0.51&0.74&0.60&\textbf{0.77}\contour{green}{($\Uparrow$}\contour{green}{51$\%$)}&\textbf{0.80}\contour{green}{($\Uparrow$}\contour{green}{8$\%$)}&\textbf{0.78}\contour{green}{($\Uparrow$}\contour{green}{30$\%$)}\\ 
\hline
Animation&\textbf{0.86}&\textbf{0.98}&\textbf{0.91}&0.85\contour{red}{($\Downarrow$}\contour{red}{1$\%$)}&0.93\contour{red}{($\Downarrow$}\contour{red}{5$\%$)}&0.89\contour{red}{($\Downarrow$}\contour{red}{2$\%$)}\\ 
\hline
Children's&0.92&0.60&0.72&\textbf{0.94}\contour{green}{($\Uparrow$}\contour{green}{2$\%$)}&\textbf{0.72}\contour{green}{($\Uparrow$}\contour{green}{20$\%$)}&\textbf{0.82}\contour{green}{($\Uparrow$}\contour{green}{14$\%$)}\\ 
\hline
Comedy&0.88&0.71&0.79&\textbf{0.89}\contour{green}{($\Uparrow$}\contour{green}{1$\%$)}&\textbf{0.84}\contour{green}{($\Uparrow$}\contour{green}{17$\%$)}&\textbf{0.87}\contour{green}{($\Uparrow$}\contour{green}{10$\%$)}\\ 
\hline
Crime&0.32&0.55&0.40&\textbf{0.68}\contour{green}{($\Uparrow$}\contour{green}{112$\%$)}&\textbf{0.76}\contour{green}{($\Uparrow$}\contour{green}{38$\%$)}&\textbf{0.72}\contour{green}{($\Uparrow$}\contour{green}{80$\%$)}\\ 
\hline
Documentary&0.60&0.50&0.55&\textbf{0.83}\contour{green}{($\Uparrow$}\contour{green}{38$\%$)}&\textbf{0.83}\contour{green}{($\Uparrow$}\contour{green}{66$\%$)}&\textbf{0.83}\contour{green}{($\Uparrow$}\contour{green}{51$\%$)}\\ 
\hline
Drama&0.70&0.86&0.77&\textbf{0.86}\contour{green}{($\Uparrow$}\contour{green}{22$\%$)}&\textbf{0.90}\contour{green}{($\Uparrow$}\contour{green}{4$\%$)}&\textbf{0.88}\contour{green}{($\Uparrow$}\contour{green}{14$\%$)}\\ 
\hline
Fantasy&0.32&\textbf{1.00}&0.48&\textbf{1.00}\contour{green}{($\Uparrow$}\contour{green}{212$\%$)}&0.92\contour{red}{($\Downarrow$}\contour{red}{8$\%$)}&\textbf{0.96}\contour{green}{($\Uparrow$}\contour{green}{200$\%$)}\\ 
\hline
Film-Noir&\textbf{1.00}&0.50&0.67&\textbf{1.00}\contour{blue}{($=$}\contour{blue}{0$\%$)}&\textbf{1.00}\contour{green}{($\Uparrow$}\contour{green}{100$\%$)}&\textbf{1.00}\contour{green}{($\Uparrow$}\contour{green}{49$\%$)}\\ 
\hline
Horror&0.75&0.80&0.77&\textbf{1.00}\contour{green}{($\Uparrow$}\contour{green}{33$\%$)}&\textbf{0.83}\contour{green}{($\Uparrow$}\contour{green}{4$\%$)}&\textbf{0.91}\contour{green}{($\Uparrow$}\contour{green}{18$\%$)}\\ 
\hline
Musical&0.80&0.80&0.80&\textbf{0.90}\contour{green}{($\Uparrow$}\contour{green}{13$\%$)}&\textbf{0.82}\contour{green}{($\Uparrow$}\contour{green}{3$\%$)}&\textbf{0.86}\contour{green}{($\Uparrow$}\contour{green}{8$\%$)}\\ 
\hline
Mystery&0.46&0.70&0.56&\textbf{0.92}\contour{green}{($\Uparrow$}\contour{green}{100$\%$)}&\textbf{0.75}\contour{green}{($=$}\contour{green}{7$\%$)}&\textbf{0.83}\contour{green}{($\Uparrow$}\contour{green}{48$\%$)}\\ 
\hline
Romance&0.59&\textbf{0.70}&0.64&\textbf{0.75}\contour{green}{($\Uparrow$}\contour{green}{27$\%$)}&\textbf{0.70}\contour{blue}{($=$}\contour{blue}{0$\%$)}&\textbf{0.73}\contour{green}{($\Uparrow$}\contour{green}{11$\%$)}\\ 
\hline
Sci-Fi&\textbf{0.88}&0.82&\textbf{0.85}&0.75\contour{red}{($\Downarrow$}\contour{red}{15$\%$)}&\textbf{0.94}\contour{green}{($\Uparrow$}\contour{green}{15$\%$)}&0.83\contour{red}{($\Downarrow$}\contour{red}{2$\%$)}\\ 
\hline
Thriller&0.76&\textbf{0.74}&0.75&\textbf{0.78}\contour{green}{($\Uparrow$}\contour{green}{2$\%$)}&0.72\contour{red}{($\Downarrow$}\contour{red}{4$\%$)}&\textbf{0.76}\contour{green}{($\Uparrow$}\contour{green}{2$\%$)}\\ 
\hline
War&0.71&0.85&0.77&\textbf{0.81}\contour{green}{($\Uparrow$}\contour{green}{14$\%$)}&\textbf{0.91}\contour{green}{($\Uparrow$}\contour{green}{7$\%$)}&\textbf{0.85}\contour{green}{($\Uparrow$}\contour{green}{10$\%$)}\\ 
\hline
Western&0.62&\textbf{1.00}&0.77&\textbf{0.71}\contour{green}{($\Uparrow$}\contour{green}{15$\%$)}&\textbf{1.00}\contour{blue}{($=$}\contour{blue}{0$\%$)}&\textbf{0.83}\contour{green}{($\Uparrow$}\contour{green}{8$\%$)}\\ 
\hline

\end{tabular}%
}
\caption{Performance of two large language models on the Movielens-100K dataset, across 18 distinct movie genres under few shot settings. The percentage variations in Precision, Recall, and F1-Score metrics, as illustrated in the fine-tuned version of ChatGPT, are compared to the baseline ChatGPT (gpt-3.5-turbo).}\label{ablation_ML_2}
\end{table}

\begin{table}
\centering
\renewcommand{\arraystretch}{1.5}
\resizebox{\columnwidth}{!}{%
\begin{tabular}{|c|c|c|c|c|c|c|}
\hline
\multirow{2}{*}{\textbf{Average Metrics}} & \multicolumn{3}{c|}{\textbf{gpt-3.5-turbo}} & \multicolumn{3}{c|}{\textbf{gpt-3.5-turbo-fine-tuned}}\\
     & Precision&Recall&F1-score& Precision&Recall&F1-score\\
\hline
Micro Avg&0.64&0.74&0.69&\textbf{0.83}\contour{green}{($\Uparrow$}\contour{green}{31$\%$)}&\textbf{0.81}\contour{green}{($\Uparrow$}\contour{green}{10$\%$)}&\textbf{0.82}\contour{green}{($\Uparrow$}\contour{green}{30$\%$)}\\ 
\hline
Macro Avg&0.69&0.71&0.67&\textbf{0.83}\contour{green}{($\Uparrow$}\contour{green}{28$\%$)}&\textbf{0.81}\contour{green}{($\Uparrow$}\contour{green}{14$\%$)}&\textbf{0.81}\contour{green}{($\Uparrow$}\contour{green}{26$\%$)}\\ 
\hline
Weighted Avg&0.71&0.73&0.71&\textbf{0.83}\contour{green}{($\Uparrow$}\contour{green}{19$\%$)}&\textbf{0.81}\contour{green}{($\Uparrow$}\contour{green}{10$\%$)}&\textbf{0.82}\contour{green}{($\Uparrow$}\contour{green}{18$\%$)}\\ 
\hline
Samples Avg&0.68&0.77&0.68&\textbf{0.85}\contour{green}{($\Uparrow$}\contour{green}{29$\%$)}&\textbf{0.83}\contour{green}{($\Uparrow$}\contour{green}{9$\%$)}&\textbf{0.82}\contour{green}{($\Uparrow$}\contour{green}{20$\%$)}\\ 
\hline
\end{tabular}%
}
\caption{Performance of two large language models on the Movielens-100K dataset under few shot setting, across average metrics. The percentage variations in Precision, Recall, and F1-Score metrics, as illustrated in the fine-tuned version of ChatGPT, are compared to the baseline ChatGPT (gpt-3.5-turbo).}\label{ablation_ML_3}
\end{table}


\begin{table}

\renewcommand{\arraystretch}{1.5}
\resizebox{\columnwidth}{!}{%
\begin{tabular}{|c|c|c|c|c|c|c|}
\hline
\multirow{2}{*}{\textbf{Genre}}  & \multicolumn{3}{c|}{\textbf{gpt-3.5-turbo(only subtitle)}} & \multicolumn{3}{c|}{\textbf{gpt-3.5-turbo(subtitle+poster info.)}}\\
     & Precision&Recall&F1-score& Precision&Recall&F1-score\\
\hline
Action&\textbf{0.82}&0.55&0.66&0.72\contour{red}{($\Downarrow$}\contour{red}{13$\%$)}&\textbf{0.76}\contour{green}{($\Uparrow$}\contour{green}{38$\%$)}&\textbf{0.74}\contour{green}{($\Uparrow$}\contour{green}{12$\%$)}\\ 
\hline
Adventure&\textbf{0.49}&\textbf{0.74}&\textbf{0.59}&0.47\contour{red}{($\Downarrow$}\contour{red}{4$\%$)}&0.69\contour{red}{($\Downarrow$}\contour{red}{6$\%$)}&0.56\contour{red}{($\Downarrow$}\contour{red}{5$\%$)}\\ 
\hline
Animation&\textbf{0.86}&\textbf{1.00}&\textbf{0.92}&0.82\contour{red}{($\Downarrow$}\contour{red}{4$\%$)}&0.85\contour{red}{$\Downarrow$}\contour{red}{15$\%$)}&0.84\contour{red}{($\Downarrow$}\contour{red}{8$\%$)}\\ 
\hline
Children's&\textbf{0.91}&0.50&0.65&0.87\contour{red}{($\Downarrow$}\contour{red}{4$\%$)}&\textbf{0.58}\contour{green}{($\Uparrow$}\contour{green}{16$\%$)}&\textbf{0.70}\contour{green}{($\Uparrow$}\contour{green}{8$\%$)}\\ 
\hline
Comedy&\textbf{0.88}&0.72&0.79&0.80\contour{red}{($\Downarrow$}\contour{red}{9$\%$)}&\textbf{0.82}\contour{green}{($\Uparrow$}\contour{green}{14$\%$)}&\textbf{0.81}\contour{green}{($\Uparrow$}\contour{green}{1$\%$)}\\ 
\hline
Crime&0.22&0.53&0.31&\textbf{0.34}\contour{green}{($\Uparrow$}\contour{green}{54$\%$)}&\textbf{0.73}\contour{green}{($\Uparrow$}\contour{green}{38$\%$)}&\textbf{0.46}\contour{green}{($\Uparrow$}\contour{green}{48$\%$)}\\ 
\hline
Documentary&0.60&0.50&0.55&\textbf{0.79}\contour{green}{($\Uparrow$}\contour{green}{31$\%$)}&\textbf{0.68}\contour{green}{($\Uparrow$}\contour{green}{36$\%$)}&\textbf{0.73}\contour{green}{($\Uparrow$}\contour{green}{32$\%$)}\\ 
\hline
Drama&0.70&\textbf{0.86}&\textbf{0.77}&\textbf{0.72}\contour{green}{($\Uparrow$}\contour{green}{2$\%$)}&0.84\contour{red}{($\Downarrow$}\contour{red}{2$\%$)}&\textbf{0.77}\contour{blue}{($=$}\contour{blue}{0$\%$)}\\ 
\hline
Fantasy&\textbf{0.24}&\textbf{1.00}&\textbf{0.38}&0.13\contour{red}{($\Downarrow$}\contour{red}{45$\%$)}&0.71\contour{red}{($\Downarrow$}\contour{red}{29$\%$)}&0.22\contour{red}{($\Downarrow$}\contour{red}{42$\%$)}\\ 
\hline
Film-Noir&\textbf{1.00}&\textbf{0.50}&\textbf{0.67}&0.80\contour{red}{($\Downarrow$}\contour{red}{20$\%$)}&0.27\contour{red}{($\Downarrow$}\contour{red}{46$\%$)}&0.40\contour{red}{($\Downarrow$}\contour{red}{40$\%$)}\\ 
\hline
Horror&0.70&0.70&0.70&\textbf{0.71}\contour{green}{($\Uparrow$}\contour{green}{1$\%$)}&\textbf{0.83}\contour{green}{($\Uparrow$}\contour{green}{18$\%$)}&\textbf{0.77}\contour{green}{($\Uparrow$}\contour{green}{8$\%$)}\\ 
\hline
Musical&\textbf{0.80}&\textbf{0.80}&\textbf{0.80}&0.67\contour{red}{($\Downarrow$}\contour{red}{16$\%$)}&0.76\contour{red}{($\Downarrow$}\contour{red}{5$\%$)}&0.71\contour{red}{($\Downarrow$}\contour{red}{11$\%$)}\\ 
\hline
Mystery&\textbf{0.26}&\textbf{0.67}&\textbf{0.38}&0.21\contour{red}{($\Downarrow$}\contour{red}{19$\%$)}&0.54\contour{red}{($\Downarrow$}\contour{red}{19$\%$)}&0.30\contour{red}{($\Downarrow$}\contour{red}{19$\%$)}\\ 
\hline
Romance&\textbf{0.52}&\textbf{0.76}&\textbf{0.62}&0.48\contour{red}{($\Downarrow$}\contour{red}{7$\%$)}&0.73\contour{red}{($\Downarrow$}\contour{red}{3$\%$)}&0.58\contour{red}{($\Downarrow$}\contour{red}{6$\%$)}\\ 
\hline
Sci-Fi&\textbf{0.88}&\textbf{0.82}&\textbf{0.85}&0.81\contour{red}{($\Downarrow$}\contour{red}{7$\%$)}&0.77\contour{red}{($\Downarrow$}\contour{red}{6$\%$)}&0.79\contour{red}{($\Downarrow$}\contour{red}{7$\%$)}\\ 
\hline
Thriller&\textbf{0.56}&0.74&\textbf{0.64}&0.48\contour{red}{($\Downarrow$}\contour{red}{14$\%$)}&\textbf{0.91}\contour{green}{($\Uparrow$}\contour{green}{23$\%$)}&0.63\contour{red}{($\Downarrow$}\contour{red}{1$\%$)}\\ 
\hline
War&0.50&0.45&0.48&\textbf{0.58}\contour{green}{($\Uparrow$}\contour{green}{16$\%$)}&\textbf{0.68}\contour{green}{($\Uparrow$}\contour{green}{51$\%$)}&\textbf{0.62}\contour{green}{($\Uparrow$}\contour{green}{29$\%$)}\\ 
\hline
Western&0.62&\textbf{1.00}&0.77&\textbf{0.81}\contour{green}{($\Uparrow$}\contour{green}{31$\%$)}&0.81\contour{red}{($\Downarrow$}\contour{red}{19$\%$)}&\textbf{0.81}\contour{green}{($\Uparrow$}\contour{green}{5$\%$)}\\ 
\hline

\end{tabular}%
}
\\
\caption{Performance of gpt-3.5-turbo on the Movielens-100K dataset under different prompts, across multiple genres. Here gpt-3.5-turbo(subtitle+poster info.) is the result of the integration of VLM and LLM. The percentage variation is with respect to the baseline gpt-3.5-turbo(only subtitle).}\label{ablation_ML_4}
\end{table}

\begin{table}
\renewcommand{\arraystretch}{1.5}
\resizebox{\columnwidth}{!}{%
\begin{tabular}{|c|c|c|c|c|c|c|}
\hline
\multirow{2}{*}{\textbf{Average Metrics}} & \multicolumn{3}{c|}{\textbf{gpt-3.5-turbo(only subtitle)}} & \multicolumn{3}{c|}{\textbf{gpt-3.5-turbo(subtitle+poster info.)}}\\
     & Precision&Recall&F1-score& Precision&Recall&F1-score\\
\hline
Micro Avg&\textbf{0.62}&0.73&0.67&0.60\contour{red}{($\Downarrow$}\contour{red}{3$\%$)}&\textbf{0.78}\contour{green}{($\Uparrow$}\contour{green}{7$\%$)}&\textbf{0.68}\contour{green}{($\Uparrow$}\contour{green}{1$\%$)}\\ 
\hline
Macro Avg&\textbf{0.64}&0.71&\textbf{0.64}&0.62\contour{red}{($\Downarrow$}\contour{red}{3$\%$)}&\textbf{0.72}\contour{green}{($\Uparrow$}\contour{green}{1$\%$)}&\textbf{0.64}\contour{blue}{($=$}\contour{blue}{0$\%$)}\\ 
\hline
Weighted Avg&\textbf{0.69}&0.73&0.69&0.66\contour{red}{($\Downarrow$}\contour{red}{4$\%$)}&\textbf{0.78}\contour{green}{($\Uparrow$}\contour{green}{7$\%$)}&\textbf{0.70}\contour{green}{($\Uparrow$}\contour{green}{1$\%$)}\\ 
\hline
Samples Avg&\textbf{0.66}&0.75&0.67&0.65\contour{red}{($\Downarrow$}\contour{red}{1$\%$)}&\textbf{0.82}\contour{green}{($\Uparrow$}\contour{green}{9$\%$)}&\textbf{0.69}\contour{green}{($\Uparrow$}\contour{green}{3$\%$)}\\ 
\hline
\end{tabular}%
}
\caption{Performance of gpt-3.5-turbo on the Movielens-100K dataset under different prompts, across average metrics. The percentage variation is with respect to the baseline gpt-3.5-turbo(only subtitle).}\label{ablation_ML_5}
\end{table}


\begin{figure*}
\subfloat[Scenario1]{\includegraphics[width = 3in]{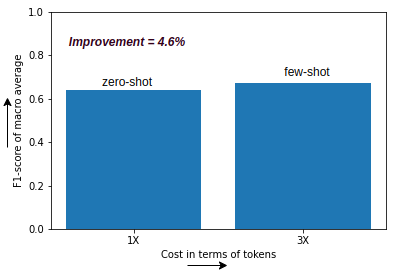}} 
\quad\quad\quad
\subfloat[Scenario2]{\includegraphics[width = 3in]{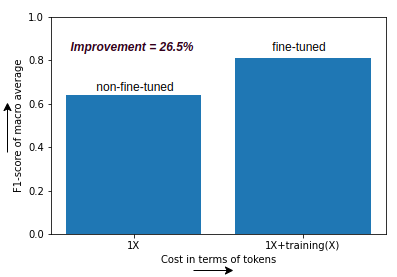}}
\caption{Cost analysis of ChatGPT under different scenarios.}
\label{example_4}
\end{figure*}

\subsection{RQ2: How does the LLMs-based genre prediction system compare with the traditional classifier method?}

In this study, we evaluated several machine-learning approaches for multi-label genre prediction on the MovieLens-100K dataset. Three traditional classification algorithms were tested: K-nearest neighbors (KNN) \cite{taunk2019brief}, logistic regression \cite{wright1995logistic}, and support vector machines (SVM) \cite{noble2006support}. The models were trained on incrementally smaller proportions of the full training set, ranging from 100\% to just 1\%. The input features to these classifiers were the movie's trailer's subtitles, whose embedding of dimensions $\mathbb{R}^{1\times768}$ were extracted from the pre-trained model SBERT\cite{reimers2019sentence}. 

\begin{figure*}
\subfloat[Precision Comparison]{\includegraphics[width = 3in]{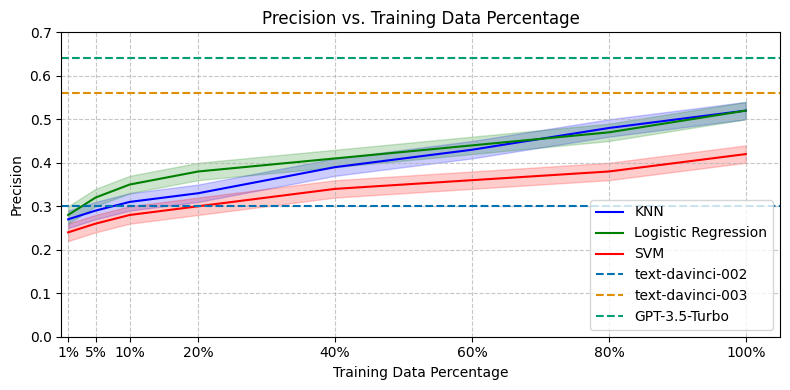}} 
\quad
\subfloat[Recall Comparison]{\includegraphics[width = 3in]{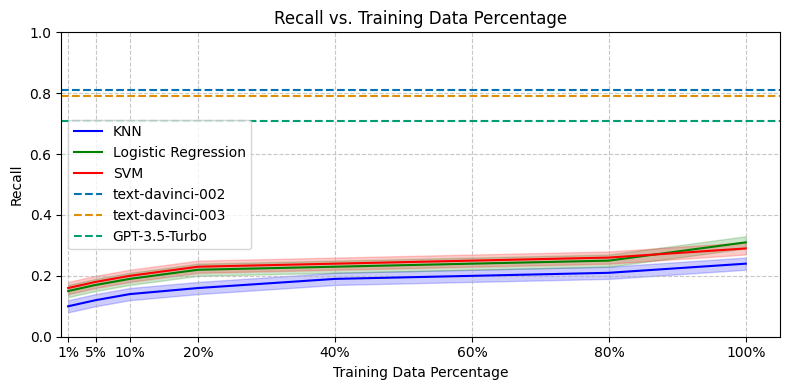}}
\\
\centering\subfloat[F1-score Comparison]{\includegraphics[width = 3in]{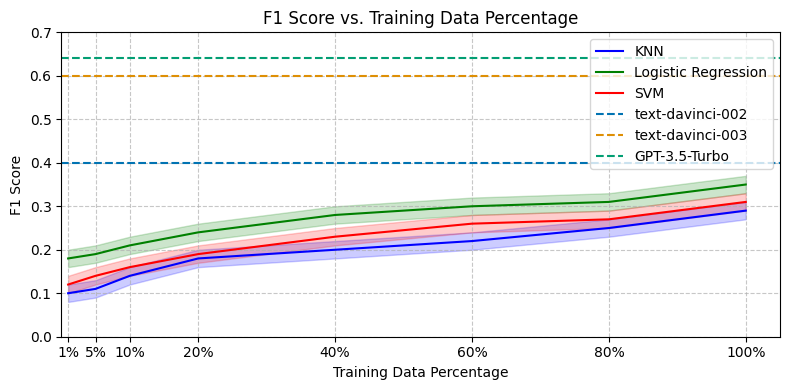}}
\caption{Comparison of three classification models with different LLMs under zero-shot settings. The shaded area indicates the 95\% confidence interval of t-distribution under 6 different experiments.}
\label{example_2}
\end{figure*}

As seen in Figure \ref{example_2}, LLMs achieved stronger performance, with gpt-3.5-turbo reaching the best macro F1-score of $0.64$, exceeding the traditional classifiers. The LLMs also showed greater robustness to limited data, maintaining decent performance even with zero training examples. Among the classifiers, logistic regression was the most robust, outperforming KNN and SVM given the same training set size. Overall, the LLMs demonstrated superior genre prediction capabilities on this movie dataset, highlighting their usefulness for multi-label genre classification tasks.

\subsection{RQ3: How much is the trade-off between the cost of gpt-3.5-turbo and its improvement?}

In order to analyze the $Cost_{chatgpt}$. we considered two scenarios, one is the improvement of a few shot settings over zero shot settings, and the second is fine-tuned over a non-fine-tuned version. Here cost will be analyzed in terms of tokens needed for the input since the output tokens will be the same in every setting, so it will not be part of our evaluation. Mathematically, we have represented as:
\begin{equation}
    Cost_{chatgpt} = N_{tokens}*C_{token}
\end{equation}
where $N_{tokens}$ is the number of input tokens, $C_{token}$ is the cost of each token. In order to simplify it, we have further considered $N_{tokens}$ in terms of $Scaling Factor$. 

We have plotted a graph in each scenario of improvement in terms of the F1-score of macro average vs cost in terms of tokens. 
\subsubsection{few-shot vs zero-shot}
Our experimental study demonstrated in Table \ref{ablation_ML_2} and \ref{ablation_ML_3} involved two-shot learning. Since no training of tokens is involved in this scenario, therefore $C_{token}$ will remain the same for both settings. Let $N_{tokens}$  considered for zero-shot be denoted as $1X$. Here, $1X$ represents the scaling factor. For the few-shot, it is denoted as $3X$, as the input tokens needed thrice the zero-shot because of the additional two examples provided for learning as shown in Figure \ref{example_3}(b). As shown in Figure \ref{example_4}(a), the graph clearly states that the improvement(\textbf{4.6\%}) has not been significant in terms of F1-score as the cost increases. 

\subsubsection{fine-tuned vs non-fine-tuned}
Fine-tuning ChatGPT involved the cost of training the input tokens. As stated in \href{https://openai.com/pricing}{openai-pricing}, it involves an additional \textbf{six times} the cost of training the tokens. For the non-fine-tuned, $N_{tokens}$ can be considered as $1X$ because it is under a zero-shot setting, but for the fine-tuned version, it will have an additional $C_{token}$ denoted here as $training(X)$. Figure \ref{example_4}(b) shows that the additional training cost has helped in improving the performance of genre prediction task significantly(\textbf{26.5\%}) compared to the performance improvement as shown in Figure \ref{example_4}(a).

\textbf{Overall, we recommend that the zero-shot settings have proved to be beneficial along with the fine-tuning for the genre prediction task considering the cost involved in accessing ChatGPT.}

\subsection{RQ4: How does the number of prompt shots affect the performance of gpt-3.5-turbo?}

\begin{figure}
    \includegraphics[width=0.5\textwidth]{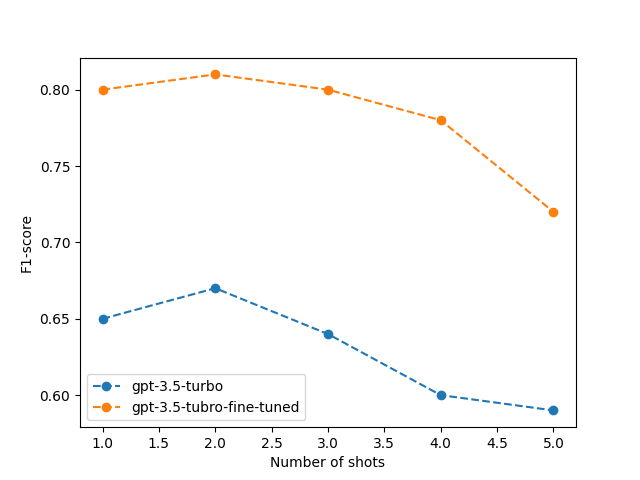}
    \caption{Effect of number of shots on LLMs for the Movielens-100K Dataset}
    \label{shot}
\end{figure}

Past studies\cite{wu2024personalized,xu2024prompting} have shown that the number of examples $E$ affects in-context learning. Here, we have an analysis of performance in terms of the F1-score of the macro average vs number of prompt shots for gpt-3.5-turbo and gpt-3.5-turbo fine-tuned. The number of prompt shots varies from 1 to 5. In Figure \ref{shot}, it can be seen that after 2 examples, performance started to decrease, which can be attributed to the fact that it might have added noise, causing LLMs to learn unwanted patterns. For our task, the optimal value of the number of prompt shots turned out to be \textbf{two}.

\section{Integration of VLM and LLM}
The extracted posters have been prompted by llava, and responses were utilized further in prompting LLMs. The extracted information from the movie poster encompasses details such as \textit{the title and typography, a vibrant and cohesive color palette, carefully chosen images and artwork providing a glimpse into the film's themes, highlighted main actors, actresses, and director, prominently displayed movie ratings and logos, strategically incorporated symbols and icons, and finally, taglines and catchphrases contributing to the overall narrative.} 

\begin{figure*}
    \centering
    \includegraphics[width=0.8\textwidth]{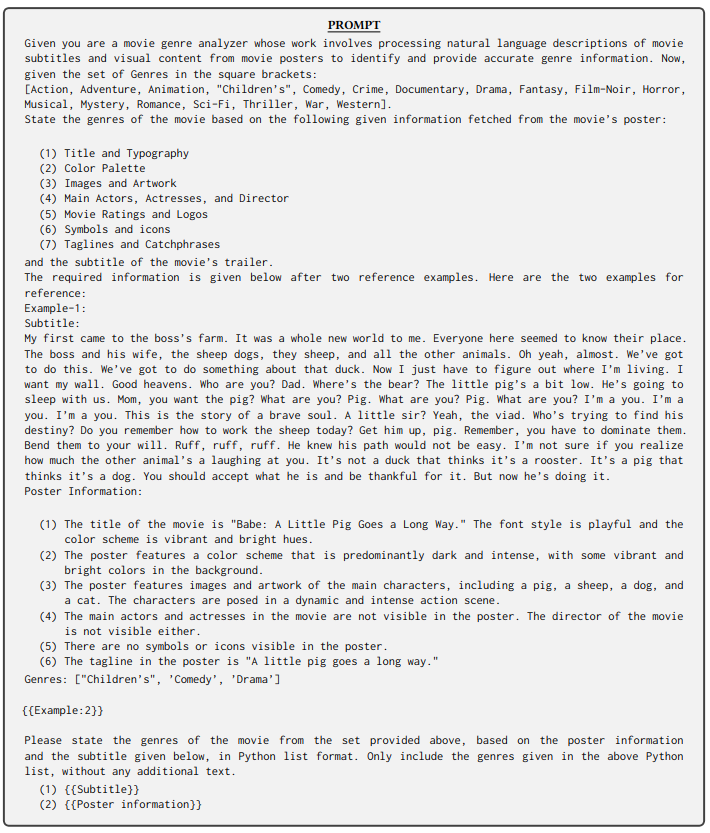}
    \caption{Prompt}
    \label{appendix1}
\end{figure*}

For this experiment, we have considered gpt-3.5-turbo for predicting the movie's genre once poster information is gathered from VLM.
The prompt given to gpt-3.5-turbo after collecting the poster information from VLM llava is shown in Figure \ref{appendix1}. Firstly, instruction is given to the LLM, following which 2 sample examples were provided, and then the movie subtitle and its poster information for which genre needs to be predicted. The experimental result is shown in Table \ref{ablation_ML_4} and \ref{ablation_ML_5}. It can be seen that the performance for genres such as \textbf{action, children's, comedy, crime, documentary, horror, war, and western} has improved, but for others, there is a decrease in performance. The genres like action, war, crime, and horror often have intense colors and dynamic visuals, making them more recognizable. In Table \ref{ablation_ML_5}, F1-score has improved in all the average metrics due to getting high recall. However, in most of the cases, precision has declined, and its case study is in the section \ref{B}.

\section{Case study for low precision}\label{B}
\begin{figure}
    \includegraphics[width=0.5\textwidth]{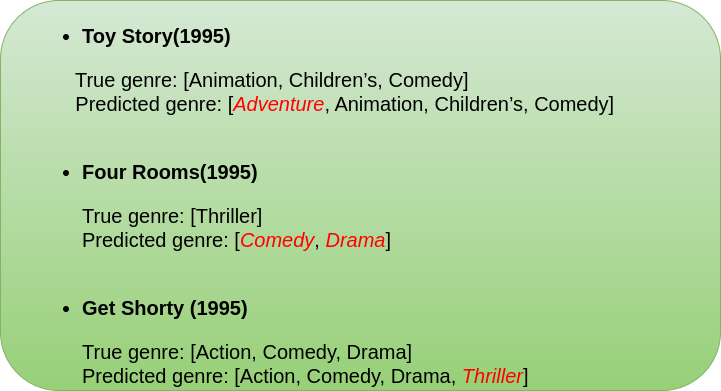}
    \caption{Case study for low precision}
    \label{casestudy}
\end{figure}

During the in-depth analysis of predicted genres obtained through the integration of VLM and LLM, it came to light that the genres assigned to movies in the Movielens-100K dataset are either limited or, in some instances, inappropriate. Illustrated in Figure \ref{casestudy}, three movies—namely \textbf{Toy Story (1995), Four Rooms (1995), and Get Shortly (1995)}—are examined.

Take the case of Toy Story, where the true genres are listed as Animation, Children's, and Comedy. However, the prediction also includes the Adventure genre, which, upon verification through IMDb and other sources, is found to be valid for this movie. Despite this relevancy, the precision for the Adventure genre diminishes during evaluation due to it being labeled as a false positive.

Similarly, for the movie Four Rooms, the genre "Thriller" is deemed inappropriate, a fact confirmed through this \href{https://www.imdb.com/title/tt0113101/}{link}. Nonetheless, the information predicted by LLM proves accurate for this movie.

This study delves into the reasons behind the low precision observed while predicting genres for movies in the Movielens dataset.

\section{Conclusion and Future Works}
In this paper, we preliminary evaluated multi-label genre prediction for the Movielens-100K dataset using four large language models, namely text-davinci-002, text-davinci-003, gpt-3.5-turbo and gpt-3.5-turbo-fine-tuned under zero-shot settings. However, under few-shot settings, we restricted it to gpt-3.5-turbo and gpt-3.5-turbo-fine-tuned. Our results indicated that fine-tuning ChatGPT under zero-shot or few-shot improves the overall performance remarkably, and it is much better than other LLMs and traditional classifiers such as logistic regression, KNN, and SVM when SBERT based embeddings are given as input.  

We have extended our study by integrating VLM to extract the movie's poster information and LLM to incorporate additional information generated by VLM. However, our experimental findings suggested that further enhancements could be achieved by fine-tuning VLM to generate more relevant and task-specific information.   


Our preliminary findings validate the efficacy of fine-tuning ChatGPT for multi-label genre prediction and set the stage for a more comprehensive investigation into its capabilities. The promising results obtained in this study underscore the significance of further research and experimentation with ChatGPT, paving the way for a deeper understanding of its implications in recommendation systems and beyond.

\bibliographystyle{IEEEtran}
\bibliography{sample}


 





\end{document}